\documentclass[10pt,a4paper,final]{iopart}

\usepackage{iopams}  
\usepackage{graphicx}
 \usepackage{nopageno}
 \usepackage[colorinlistoftodos,textsize=tiny]{todonotes}

\begin{document}
\title[]{Transferability of Operational Status Classification Models Among Different Wind Turbine Types}

\author{Z. Trstanova$^1$, A. Martinsson$^1$, C. Matthews$^1$, S. Jimenez$^2$, B. Leimkuhler$^1$, T. Van Delft$^2$, M. Wilkinson$^2$}
\address{$^1$University of Edinburgh, Edinburgh, United Kingdom}
\address{$^2$DNV GL, Oslo, Norway}
\eads{\mailto{zofia.trstanova@ed.ac.uk}, \mailto{michael.wilkinson@dnvgl.com}}

\begin{abstract}
 A detailed understanding of wind turbine performance status classification can improve operations and maintenance in the wind energy industry. Due to different engineering properties of wind turbines, the standard supervised learning models used for classification do not generalize across data sets obtained from different wind sites. We propose two methods to deal with the transferability of the trained models: first, data normalization 
 in the form of power curve alignment, and second, a robust method based on convolutional neural networks and feature-space extension. We demonstrate the success of our methods on real-world data sets with industrial applications.
 \end{abstract}

\vspace{2pc}

\section{Introduction}


Classification of operational status is an important step for performance analysis of wind farms from data of SCADA (Supervisory Control and Data Acquisition) type. The main goal of the analysis is future energy production assessment and the identification of performance issues. The analysis of a typical wind farm may consist of hundreds of millions of SCADA records. Until now, classification of under-performance and different operational states has been performed by a human analyst, automatization of this process can save many working hours and provide additional insight into wind turbine operations.
Typically, companies possess historical data from various farms where the power generation status has already been analyzed. Understanding the operational status can explain under-performance, for example by indicating when the turbine may require maintenance. This under-performance may be a consequence of power derating, blade icing, or other conditions. In order to automate this task, we use the knowledge of these under-performing operations within a supervised learning model, which is then employed to classify unlabeled data from various turbine types and farms.

Machine learning has previously been applied to wind turbine SCADA data: (i) for example, time-series regression setting for power prediction~\cite{Clifton2013,Sapronova2016,cnn2017}, and (ii) for fault classification~\cite{berkley2016}. In this paper, we are interested in multi-class classification of the operational status of power generation for post-construction operational analysis. In a previous article, a classifier was applied to wind turbine fault analysis~\cite{berkley2016} using support vector machines. In that work, multiple
binary classifiers were trained on separate test sets for each class. In real-world problems, the test set contains multiple classes, which makes such an approach difficult to implement in the industrial setting. Another challenge which impedes the industrial application of trained models among different data sets, is the variation between wind turbines: turbine manufacturer, model, geographic location, etc.~\cite{gill2012wind,berkley2016,uluyol2011power}.
This results in different nacelle anemometer power curves (various cut-in and cut-out wind speeds, rated powers, etc., see  Figure~\ref{baselines}). One of the main challenges in developing a robust method is, therefore, to determine how to handle the issue of generalization of the machine learning model between different farms. 

\begin{figure}
    \centering
    \includegraphics[scale=0.25]{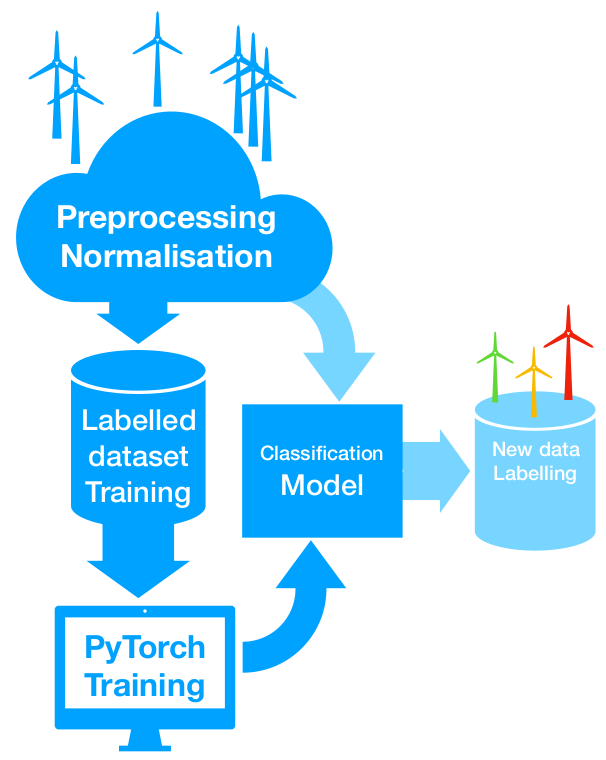}
    \caption{Diagram of the acwind software system for automatic classification of wind turbine performance.}
    \label{fig:acwindflow}
\end{figure}

To handle the transferability issue of classification of the operational state, we propose two solutions providing a robust method based on data normalization and feature-space extension using machine learning based on convolutional neural networks (CNN).  We incorporate these techniques into a python software package ``acwind'' which is planned for release in open source mode (see Figure \ref{fig:acwindflow} for a diagram illustrating its functionality).




This paper is organized as follows: In Section~\ref{data}, we describe the data sets 
and the classification challenge. In Section~\ref{methods}, we introduce baseline normalization in the preprocessing of industrial data sets and the proposed classification methods. Finally, in Section~\ref{results} we demonstrate the results of our method on real-world data.

\section{Data sets{\label{data}}}

\begin{figure}[htb!]
\begin{center}
 \includegraphics[width=0.5\textwidth]{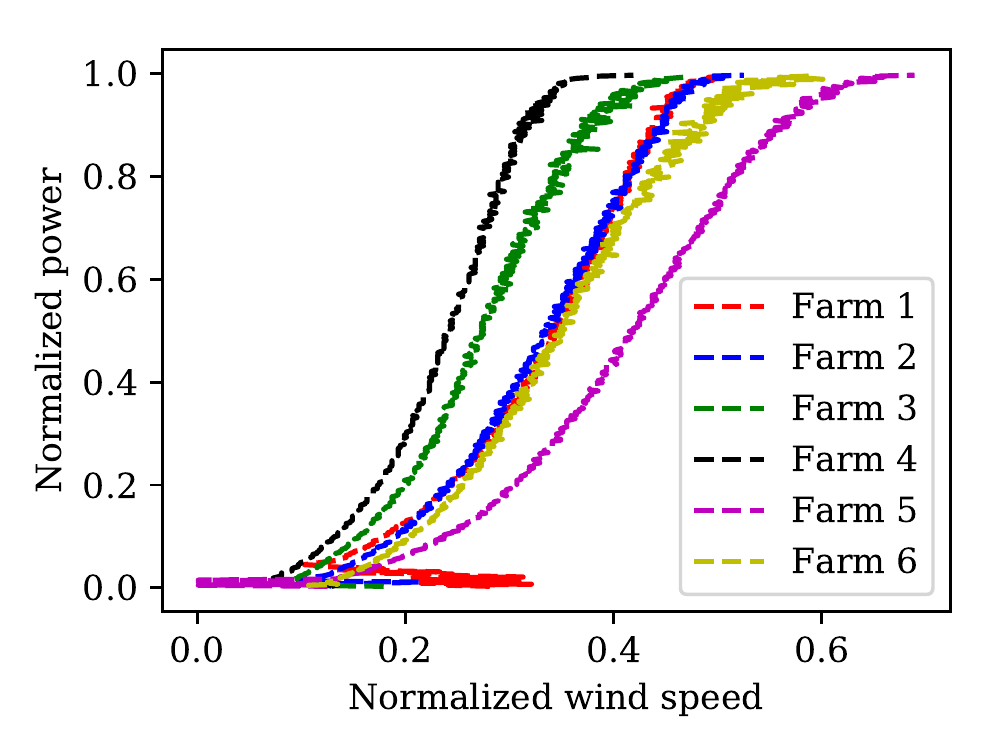}
 \caption{\label{baselines} Characteristic power curve of data from six wind turbine farms, learned from the historical time-series of wind speed and power signals.}
 \end{center}
\end{figure}

We use historical SCADA data provided by DNV GL\footnote{www.dnvgl.com}. These standard operational 10-minute SCADA   contain channels found in a typical modern wind turbine (e.g. wind speed, power, pitch, rotor speed,  ambient temperature, and others). Each data point has been manually labeled as representing either normal operation or abnormal with a particular status type (such as \textit{anemometer icing, unavailability, partial unavailability, derating, other performance issues}, for example). By a data set, we mean a collection of historical SCADA data from one wind farm. More precisely, these data sets relate to time series data for six farms (i.e. collections of wind turbines of the same type and location), with each time point classed as  ``normal", $C_i$ (with $i=1, \ldots 4$) or ``other". We denote by ``other" a class which contains the remaining potential classes of the data set. The data sets contain time series over periods between 18 and 28 months, and with between five and fifteen individual turbines in each farm. Pairs of farms $(1,2), \, (3,4)$ and $(5,6)$ have similar classes but also  different power curves (Figure~\ref{baselines}). Note that the data sets include neither warning nor status data, although this information might have been used during the manual labeling process. As part of the preprocessing, the data sets have been anonymized by multiplication by a constant factor and min-max normalized to values in $[0, 1]$.


The wind speed power curve is a useful metric for identifying periods of relative under-performance of a wind turbine, and it can be learned from the data~\cite{gonzalez2017use,park2014development}.  To deal with the deviations among various data sets, we use a statistical method to learn the reference power curve for each turbine type from historical data. This is possible under the assumption that the turbine operates normally most of the time (in our data sets, around $80\%$ of points were labeled as normal). The power curve baseline can then be obtained by two-dimensional histogram analysis of the operational data, providing for each discrete value of power the value of the corresponding wind speed with maximum frequency.
 Figure~\ref{baselines} shows the baselines of six farms inferred from the operational data. 

\section{Methods \label{methods}}

\subsection{Baseline normalization}
The goal is to train a supervised learning model on a reference data set, for which various operational states have been identified. As mentioned above, the main challenge is that different farms have a different underlying distribution of SCADA signal values, leading to very different normal behavior described by the baseline, see Figure~\ref{baselines}. The aim of this section is to describe a method which allows for the generalization of the classification model to data sets with very different baselines. 

The main idea is to preprocess each data set associated with a particular wind farm in such a way that the baselines among the farms are superimposed.
In order to map the baseline of one data set onto another "reference" farm, we initially scale the power signal so that all entries lie between 0 and 1. We then apply a linear transformation to the wind speed signal, by multiplying it by a constant $\alpha$ and shifting by a constant $\beta$. The values of $\alpha$ and $\beta$ are chosen to minimize the L1 norm (integrated absolute value) of the difference between the histogrammed distributions of wind speed and power, relative to those of the target wind farm. Standard optimization techniques can be used to achieve this minimization efficiently.

\begin{figure}[htb!]
\begin{center}
 \includegraphics[width=1.0\textwidth]{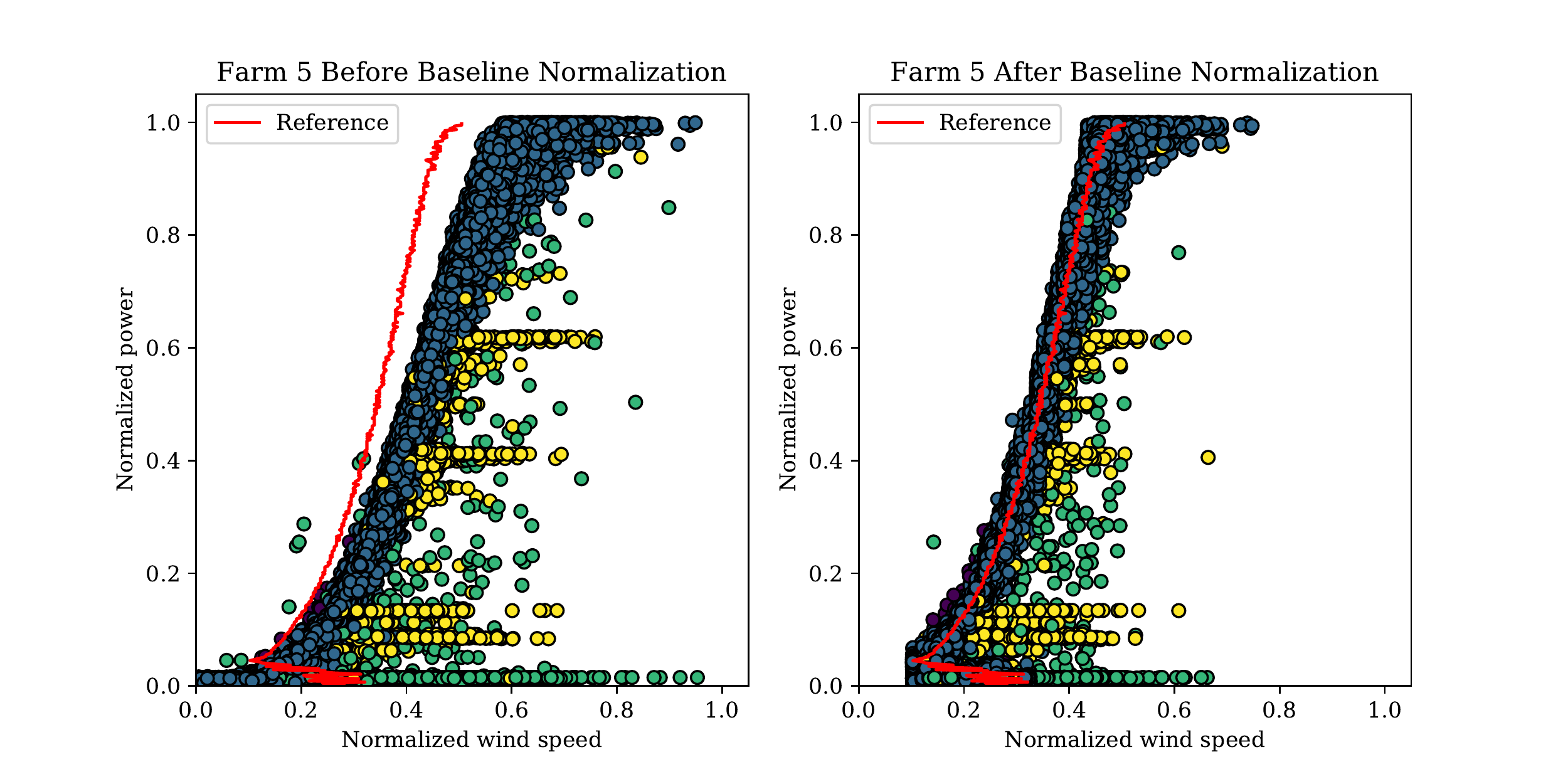}
 \caption{\label{normalization} Normalization of data set from wind farm 5 with respect to the reference baseline of wind farm 1. Individual points are colored
 by the class type provided by four labels, dark blue represents the normal class.}
 \end{center}
 \end{figure}

This transformation only requires finding two parameters and only need be done once. Even though nonlinear transformations can be applied to relate the two distributions, we have found that this simple linear transformation is sufficient to improve the generalization across training. Figure~\ref{normalization} illustrates such baseline normalization.

\subsection{Model accuracy}

We consider a data set $\{(x_1, y_1), \ldots, (x_N, y_N)\}$ containing $N$ data points $x_i$ with associated class labels $y_i$, where $y_i\in\{1,2,\ldots,n\}$. Each data point $x_i$ is in $\mathbb{R}^p$ for $p$ available signals. A supervised learning model parameterizes a function $f_{\theta}(\cdot)$, for parameter vector $\theta$,  such that $f_{\theta}(X)\approx Y$~\cite{introstat}.
The accuracy of a classifier can be estimated by the average over the correct predictions on the training data set,
\begin{equation}
\label{score}
\frac{1}{N}\sum_{i=1}^{N}\delta(y_i - f(x_i)),
\end{equation}
where $N$ is the number of data points in the set,  and $\delta$ represents a Dirac delta function. 

In order to prevent overfitting on the training data set, we measure the average score (\ref{score}) over the test set.  Due to a class imbalance problem, we estimate the average accuracy per class in the test set. 



\subsection{KNN and feed-forward neural network}

\textit{K}-nearest neigbors(KNN) algorithm~\cite{introstat} is a non-parametric method, which, for a particular point $x$ estimates the conditional probability of belonging to a class $C_j$ as the fraction of k-nearest neighbors of $x$ that belong to the class $C_j, j=1, \ldots M$. We use the Euclidean distance to define the set of \textit{k}-nearest neighbors $S_K(x)$.  More precisely, the probability of a point $x$ belonging to the class $C_j$ is
\[
\mathbb{P}(y = C_j\mid X = x) = \frac{1}{K}\sum_{i\in S_K(x)} \delta(y_i - C_j).
\]
The point $x$ is then classified according to Bayes formula by the class with the largest estimated probability.

Feed-forward neural networks are nonlinear models with specific parameters and layered structure of nodes/neurons, historically motivated by neural science. The parameters of each layer $l$ consist of weights and biases $\theta_l = (w_l, b_l)$ which are associated to connections between the layers.  More precisely, the output of layer $\ell$ is given by $\sigma(w_{\ell}^Tx + b_{\ell})$ where $x$ is either the input from the previous layer, or (for $\ell=1$) the data itself, and $\sigma$ is an activation function. The loss function for a classification task is the cross-entropy, which is based on maximum likelihood estimation of parameters, aiming at maximizing the probability to observe the labeled data set $(x, y)$. It is possible to access the inferred probabilities of the classifier, however, they might be misleading in the estimation of model uncertainty and Bayesian approaches based on a sampling of the posterior should be favored for this task.  For more details on neural networks, see for example~\cite{murphy2013machine}. One of the main advantages of neural networks is that, due to their structure and the backpropagation algorithm (based on the chain rule), it is possible to efficiently parameterize the  networks even for many parameters and for large data sets. Libraries as PyTorch~\cite{paszke2017automatic} or Tensorflow~\cite{abadi2016tensorflow} provide a framework for efficient handling of distributed training on GPUs.

%

\subsection{Convolutional neural networks}

Convolutional neural networks (CNN's) are feed-forward neural networks that have had great success in image and speech recognition~\cite{Lecun2015}. In addition to the fully connected layers, they incorporate  convolutional and pooling layers, allowing for feature extraction from a data set, e.g an image, reducing the size of the representation and creating shift invariance. 

\begin{figure}[h!]
 \centering
 \includegraphics[width=0.5\textwidth]{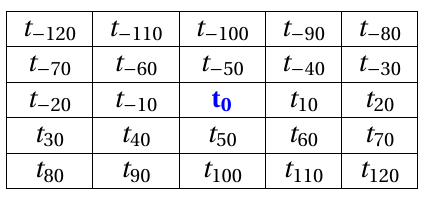}
 \caption{\label{matrix}An augmented data point from the local time series organized as a matrix: The central point determines the class label. }
\end{figure}

Time-dependence is an important factor in many of the classes. Motivated by this idea, we extend the features of each data point by the local time series. As in~\cite{cnn2017}, we first preprocess the data by arranging it into a matrix, with the principal data point (with timestamp $t_0$) located in the center. 
Figure~\ref{matrix} shows such an augmented data point, where $t_k$ indicates a point at time $t_0 + k$ minutes. 

Other options for this matrix are available, for example in the case of online classification we may wish to only consider previous data points. This could easily be used without other changes to the methodology we describe.


We use the following architecture of the feed-forward CNN: two convolutional layers (32 output signals and kernel with size 3) followed by a three-layer feed-forward neural network with ReLU activation functions. We use the cross-entropy as a cost function and we train the network using PyTorch~\cite{paszke2017automatic} (Adam optimizer, learning rate $0.01$). We are aware of the existence of neural networks specifically designed to handle time series as, for example, recurrent neural networks, however we opted to use CNNs for their translation-invariance property, and also motivated by the fact that the labeling process is mostly done by an analyst looking at feature space projections and very occasionally reviewing the time series. As we will show in the results section, extending the feature space in the local time series indeed improves the predictions.


\section{\label{results}Results and discussion}

The results described in this section were obtained using our \textit{acwind}~\cite{acwind}, which is an open source software
package for automatic classification of wind
energy analytics. The package requires
PyTorch and can be used to train CNN's to
classify the operational status of SCADA data.
We compare the methods proposed in Section~\ref{methods}: KNN or feed-forward neural network with baseline normalization and CNN. We demonstrate that a CNN does not require data normalization during preprocessing and that the model automatically generalizes among farms with various turbine types.

We do not expect high accuracy for predictions of the ``other" class since it contains a different mixture of classes in every data set and the trained model should not relate them. Class $C_2$ is a very small class, which contains points of unknown labels, also denoted as ``spurious". It is therefore difficult to learn and we do not expect to achieve high accuracies for this class, either. When the model is trained on the data set from Farm $j$, the accuracy is estimated on its test subset (splitting ratio $1/3$). We then evaluate the model on Farm $k$, which we refer to as ``test set'' in the tables below. 
%

In the first example, we preprocess
the data sets with baseline normalization and use a KNN model or a feed-forward neural network. 
More precisely, after the baseline normalization of the data sets, we parameterize/train two supervised learning models: \textit{k}-nearest neighbors (KNN) with $K=50$ neighbors and a feed-forward neural network with three hidden layers with $(12, 6, 6)$ nodes respectively and rectified linear unit (ReLU) activation functions. We use the cross-entropy as cost function. The hyperparameters were chosen by  cross-validation on various subsets of the data set.
In Table~\ref{knn 2} we show results using 3 signals. Table~\ref{fnn} shows equivalent results for a feed-forward neural network. In both cases, the normalization improves the generalization between the data sets. 
 

%
    Table~\ref{cnn 6} shows results using the CNN with four input signals: wind-speed, power, rotor-speed, and pitch. We observe an increase in average accuracy and improved generalization between different farms. In our data sets, only a few classes are present in all six data sets. For example, in the data sets used for results in Table~\ref{cnn 6}, the $C_4$ class is not present in Farm 1 and 2 and conversely, $C_3$ is not present in either Farm 3 nor 4. In this example, we replace these missing classes by inserting a new column of ``False" values. We can then train on six classes: ``normal", $C_1$, $C_2$, $C_3$, $C_4$ and ``other". The results in Table~\ref{cnn 6} show the robustness of the CNN in transferring the learning among various farms if there is enough representation of the target class in a data set.
    
    We also train a CNN on a \emph{mixed} data set, created by concatenating the data from Farms 1, 2 and 3 (after preprocessing them with the baseline normalization method).  The results shown in Table~\ref{cnn 6} demonstrate that using a mixed data set can give better generalization than using any one data set. We ran experiments using other combinations of farms for the mixed data set and found qualitatively similar results.


We grouped the data sets such that there is a representation of the specific class ($C_4$ or $C_3$) in the test set. The results from Table~\ref{cnn 1} and~\ref{cnn 2} reinforce the conclusion that, with a good representation of a particular class, CNN can learn the classes very well.

Finally, we note that the machine learning model requires consistency in labeling the data sets used for training since in the absences of this lower accuracies are observed during cross-validation. This issue cannot be resolved by increasing the size of the training data set and can only be addressed by ensuring that the training set does not contain any ambiguities of operational status interpretation.


\begin{table}
\centering
\begin{tabular}{ccc}
     Accuracy $\%$ \\
     \br
     Test Set & Farm 1 & Farm 3 \\
     \hline
Farm 1$^{\rm a}$ &  $ 99  (100 , 99 , 6 , 94 )$ & $    77  (82 , 97 , 2 , 37 )$ \\
Farm 3$^{\rm a}$ & $73  (69 , 92 , 2 , 74 )    $ & $97  (99 , 98 , 18 , 88 )$ \\
     \br
     $^{\rm a}$ Training set.\\
\end{tabular}
\caption{KNN with baseline normalization using wind-speed, power and pitch when classifying: Normal, $C_1$, $C_2$ and other}
\label{knn 2}
\end{table}

  \begin{table}
\centering
\begin{tabular}{ccc}
     Accuracy $\%$ \\
     \br
     Test Set & Farm 1 & Farm 3 \\
     \hline
     Farm 1$^{\rm a}$ &  $98  (99 , 98.6 , 31 , 90 )    $ & $72  (82 , 98 , 11 , 9 )$ \\
Farm 3$^{\rm a}$ & $69  (72 , 87 , 0 , 15 )    $ & $96  (99 , 93 , 0 , 79 )$ \\
     \br
     $^{\rm a}$ Training set.\\
\end{tabular}
\caption{Feed-forward NN with baseline normalization using wind-speed, power and pitch when classifying: Normal, $C_1$, $C_2$ and other}
\label{fnn}
\end{table}

\begin{table}[h!]
\centering
\begin{tabular}{ccc}
     Accuracy $\%$ \\
     \br
     Test Set & Farm 1 & Farm 2 \\
     \hline
     Farm 1$^{\rm a}$ & $98(99, 95, 98, 0,10)$ & $97(99,80,0,99,0,0)$ \\ 
Farm 2$^{\rm a}$ & $94(94,96,98,0,0)$ & $97(99,90,0,99,0,0)$\\ 
Farm 3$^{\rm a}$ & $60(61,0,-,94,0,9)$ & $49(53, 0, 55, 89, 0, 29)$\\ 
Farm 4$^{\rm a}$ & $81(87, 0, -, 94, 0, 0)$ &    $91(100, 0, 29, 91, 0, 0)$\\ 
Mixed$^{\rm a}$ & $98(99, 94, -, 98, 29, 0)$ &    $98(99, 85, -, 99, 27, 0)$ \\ 
     \br
    Test Set & Farm 3 & Farm 4 \\
    \hline
         Farm 1$^{\rm a}$ & $85(99,-, 0,96,0,0)$ & $94(99,-,0,95,0,0)$ \\ 
Farm 2$^{\rm a}$ &    $85 (99,-,0,96,0,0) $ & $    94(100,-,0,92,0,0)$ \\ 
Farm 3$^{\rm a}$& $97 (99, 90, 96, 1, 53)$ & $92(92, 90, 93, 3, 0)$ \\ 
Farm 4$^{\rm a}$& $94(100, 70, 95, 0, 0)$ & $99(100, 96, 96, 0,0)$ \\ 
Mixed$^{\rm a}$ & $96(98, -, 89, 96, 8,0)$ & $94(95, -, 75,97,57,0)$ \\ 
    \br
     $^{\rm a}$ Training set.\\
\end{tabular}
\caption{ CNN with wind-speed, power, rotor-speed and pitch. Classification of 6 classes: normal, $C_3$, $C_4$, $C_1$, $C_2$, other}
\label{cnn 6}
\end{table}


 \begin{table}
\centering
\begin{tabular}{ccc}
     Accuracy $\%$ \\
     \br
     Test Set & Farm 1 & Farm 2 \\
     \hline
     Farm 1$^{\rm a}$ &  $ 98  (100 , 93 , 99 , 1 , 3 ) $    & $97  (100 , 72 , 98 , 12 , 0 )$ \\
    Farm 2$^{\rm a}$ & $97  (99 , 87 , 96 , 0 , 0 )$ & $    97  (100 , 74 , 97 , 0 , 0 )$ \\
     \br
     $^{\rm a}$ Training set.\\
\end{tabular}
\caption{CNN with wind-speed, power signal, rotor-speed and pitch when classifying: Normal, $C_3$, $C_1$, $C_2$ and other}
\label{cnn 1}
\end{table}

\begin{table}
\centering
\begin{tabular}{ccc}
     Accuracy $\%$ \\
     \br
     Test Set & Farm 3 & Farm 4 \\
     \hline
     Farm 3$^{\rm a}$ &  $97  (99 , 89 , 97 , 3 , 61 )$& $    96  (97 , 88 , 93 , 0 , 1 )$ \\
     Farm 4$^{\rm a}$ & $94  (100 , 62 , 97 , 0 , 0 )    $ & $98  (100 , 85 , 97 , 0 , 0 )$ \\
     \br
     $^{\rm a}$ Training set.\\
\end{tabular}
\caption{CNN with wind-speed, power signal, rotor-speed and pitch when classifying: Normal, $C_4$, $C_1$, $C_2$ and other}
\label{cnn 2}
\end{table}

\section{Conclusions}

We have proposed two methods to deal with the generalization of classification models among different wind turbine types. In case relatively few signals are available, data normalization can improve the generalization as demonstrated on both KNN and feed-forward neural networks. The use of a CNN can provide a more robust method, which is invariant to the baseline differences and allows for the use of feature-space extension in the form of a local time series. If the class representation within the data set is sufficient, the accuracy is correspondingly high. This suggests that historical data from many farms can be combined to create a ``mixed training set" with all the possible operational states. In order to combine SCADA data from various farms into the mixed data set, each of the data sets needs to be normalized in advance. A CNN based model provides robust predictions for the specific classes, thus providing a comprehensive, robust solution for demanding industrial applications.



%



\section*{References}
\bibliographystyle{plain}

\begin{thebibliography}{10}
	
	\bibitem{abadi2016tensorflow}
	Mart{\'{i}}n Abadi, Paul Barham, Jianmin Chen, Zhifeng Chen, Andy Davis,
	Jeffrey Dean, Matthieu Devin, Sanjay Ghemawat, Geoffrey Irving, Michael
	Isard, and Others.
	\newblock {Tensorflow: a system for large-scale machine learning.}
	\newblock In {\em OSDI}, volume~16, pages 265--283, 2016.
	
	\bibitem{Clifton2013}
	A~Clifton, L~Kilcher, J~K Lundquist, and P~Fleming.
	\newblock {Using machine learning to predict wind turbine power output}.
	\newblock {\em Environ. Res. Lett.}, 2013.
	
	\bibitem{gill2012wind}
	Simon Gill, Bruce Stephen, and Stuart Galloway.
	\newblock {Wind turbine condition assessment through power curve copula
		modeling}.
	\newblock {\em IEEE Trans. Sustain. Energy}, 3(1):94--101, 2012.
	
	\bibitem{gonzalez2017use}
	E~Gonzalez, B~Stephen, D~Infield, and J~J Melero.
	\newblock {On the use of high-frequency SCADA data for improved wind turbine
		performance monitoring}.
	\newblock In {\em J. Phys. Conf. Ser.}, volume 926, page 12009. IOP Publishing,
	2017.
	
	\bibitem{introstat}
	Gareth James.
	\newblock {\em {An Introduction to Statistical Learning}}.
	\newblock Springer.
	
	\bibitem{berkley2016}
	Kevin Leahy, R~Lily Hu, Ioannis~C Konstantakopoulos, Costas~J Spanos, and
	Alice~M Agogino.
	\newblock {Diagnosing Wind Turbine Faults Using Machine Learning Techniques
		Applied to Operational Data}.
	\newblock {\em Progn. Heal. Manag. (ICPHM), 2016 IEEE Int. Conf.}, 2016.
	
	\bibitem{Lecun2015}
	Yann Lecun, Yoshua Bengio, and Geoffrey Hinton.
	\newblock {Deep learning}.
	\newblock {\em Nature}, 2015.
	
	\bibitem{acwind}
	Anton Martinsson, Zofia Trstanova, and Charles Matthews.
	\newblock https://github.com/acwind-lib/acwind, 2019.
	
	\bibitem{murphy2013machine}
	Kevin~P Murphy.
	\newblock {\em {Machine learning : a probabilistic perspective}}.
	\newblock MIT Press, Cambridge, Mass. [u.a.], 2013.
	
	\bibitem{park2014development}
	Joon-Young Park, Jae-Kyung Lee, Ki-Yong Oh, and Jun-Shin Lee.
	\newblock {Development of a novel power curve monitoring method for wind
		turbines and its field tests}.
	\newblock {\em IEEE Trans. Energy Convers.}, 29(1):119--128, 2014.
	
	\bibitem{paszke2017automatic}
	Adam Paszke, Sam Gross, Soumith Chintala, Gregory Chanan, Edward Yang, Zachary
	DeVito, Zeming Lin, Alban Desmaison, Luca Antiga, and Adam Lerer.
	\newblock {Automatic differentiation in PyTorch}.
	\newblock 2017.
	
	\bibitem{Sapronova2016}
	Alla Sapronova, Meissner Catherine, and Matteo Mana.
	\newblock {Short time ahead wind power production forecast}.
	\newblock {\em J. Phys. Conf. Ser.}, 2016.
	
	\bibitem{uluyol2011power}
	Onder Uluyol, Girija Parthasarathy, Wendy Foslien, and Kyusung Kim.
	\newblock {Power curve analytic for wind turbine performance monitoring and
		prognostics}.
	\newblock In {\em Annu. Conf. Progn. Heal. Manag. Soc.}, volume~2, pages 1--8,
	2011.
	
	\bibitem{cnn2017}
	Anwen Zhu, Xiaohui Li, Zhiyong Mo, and Huaren Wu.
	\newblock {Wind Power Prediction Based on a Convolutional Neural Network}.
	\newblock {\em Int. Conf. Circuits, Devices Syst. Wind}, 2017.
	
\end{thebibliography}

\end{document}